\documentclass[runningheads]{llncs}

\usepackage{eccv}

\usepackage{eccvabbrv}
\usepackage{graphicx}
\usepackage{booktabs}
\usepackage[accsupp]{axessibility}

\usepackage{hyperref}
\hypersetup{urlbordercolor={1 1 1}}

\usepackage{orcidlink}
\usepackage{booktabs}
\usepackage{multirow}
\usepackage{graphicx}
\usepackage{xcolor}
\usepackage{amsmath,amssymb}
\usepackage{pifont}

\newcommand{\cmark}{\ding{51}}
\newcommand{\xmark}{\ding{55}}

\begin{document}

\title{VTaMo: Video-Text Alignment Model for Sign Language Translation}

\titlerunning{VTaMo}

\author{Junyi Hu\inst{1} \and
Zhewen He\inst{1} \and
Haomian Huang\inst{1} \and
Aoxiang Yang\inst{1} \and
Yi Fang\inst{1,2}\thanks{Corresponding author.}}

\authorrunning{J. Hu et al.}

\institute{New York University Abu Dhabi, UAE \and ChatSign Technology \\
\email{jh10472@nyu.edu}, {yf23@nyu.edu}}

\maketitle

\begin{abstract}
Sign language translation (SLT) converts continuous sign videos into spoken language text.
Gloss-free approaches leverage pre-trained visual encoders and language models but rely on implicit cross-modal alignment from translation supervision alone.
We present VTaMo, a framework that introduces explicit multi-granularity alignment at three levels: (1) local alignment via entropy-regularized optimal transport with a learnable null token for fine-grained frame-to-token correspondences; (2) global alignment via a learnable orthogonal transformation that calibrates embedding space geometry through Earth Mover's Distance; and (3) position-aligned contrastive learning for discriminative token-level representations.
Experiments on Phoenix-2014T, CSL-Daily, How2Sign, and OpenASL demonstrate consistent state-of-the-art performance, with ablations confirming the complementary contributions of each component.
Code is available at \url{https://github.com/junyi2005/vtamo}.
\keywords{Sign language translation \and Cross-modal alignment \and Optimal transport}
\end{abstract}

\section{Introduction}
\label{sec:intro}

\begin{figure*}[t]
    \centering
    \includegraphics[width=\textwidth]{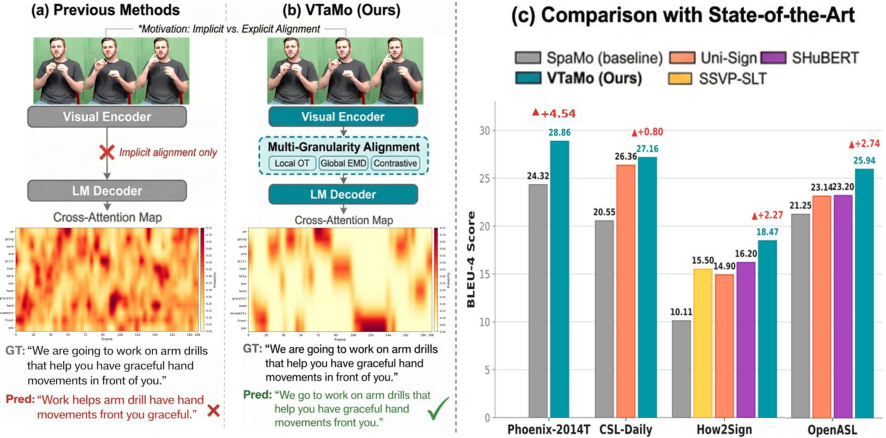}
    \caption{\textbf{Motivation and performance of VTaMo.} \textbf{(a)} Prior gloss-free sign language translation methods typically rely on \emph{implicit} cross-modal alignment learned inside the decoder attention, which can yield diffuse or mismatched cross-attention and translation errors. \textbf{(b)} VTaMo introduces \emph{explicit} multi-granularity vision--text alignment---including local OT-based token-to-frame matching, global distribution alignment, and position-aligned contrastive learning---to sharpen correspondences and improve decoding. \textbf{(c)} VTaMo achieves state-of-the-art sign language translation quality across four benchmarks (Phoenix-2014T, CSL-Daily, How2Sign, and OpenASL), compared against SpaMo~\cite{hwang2025efficient}, Uni-Sign~\cite{li2025unisign}, SHuBERT~\cite{gueuwou2025shubert}, and SSVP-SLT~\cite{rust2024towards}. Video frames in (a) and (b) are from the How2Sign dataset~\cite{duarte2021how2sign}.}
    \label{fig:intro}
\end{figure*}

Sign language translation (SLT) aims to convert continuous sign language videos into spoken language sentences, enabling more accessible communication between Deaf and hard-of-hearing signers and hearing communities~\cite{camgoz2018neural,zhou2021improving}.
Recent gloss-free systems decode text directly from visual features using large pre-trained sequence-to-sequence language models~\cite{wong2024sign2gpt,gong2024llms,chen2024factorized}, avoiding the costly gloss annotation required by gloss-based pipelines.
However, most gloss-free benchmarks provide only natural language sentences as supervision, without gloss annotations or temporal segmentation~\cite{duarte2021how2sign,shi2022open}.
The core difficulty is therefore twofold: the semantic gap between vision and text, and the unknown correspondence between the temporal order of visual signs and the token order of the target text.

Sign languages often do not follow the word order of the corresponding spoken language: a signer may express key content words first and add grammatical relations later, so the temporal progression of gestures can differ from the target word order.
As a result, visual features extracted along the video timeline are often misaligned with the text tokens that an autoregressive decoder is trained to predict.
Left unresolved, the decoder must simultaneously learn translation and implicitly discover a latent cross-modal permutation, which increases optimization difficulty and degrades accuracy, especially on large-scale benchmarks such as How2Sign and OpenASL.
Fig.~\ref{fig:intro} illustrates this gap: when alignment is left implicit, the cross-attention map can be noisy and the decoder may produce incorrect word ordering, whereas our approach makes alignment explicit.

Existing methods largely treat the visual stream as an ordered sequence and rely on decoder attention to bridge the modality gap.
ShuBERT~\cite{gueuwou2025shubert} and Uni-Sign~\cite{li2025unisign} strengthen representations through large-scale pretraining but do not explicitly model frame-to-token correspondence within a sentence, while SpaMo~\cite{hwang2025efficient} uses batch-level contrastive learning yet leaves fine-grained, sentence-internal alignment unresolved.
These limitations motivate a model that learns correspondence between visual segments and text tokens and uses it to present the decoder with a semantically ordered visual sequence.

We propose \textbf{VTaMo}, a vision--text alignment model for gloss-free SLT that explicitly aligns and reorders visual features before text generation.
Because many spoken words such as articles and prepositions have no sign-level counterpart, we align against and decode a content-word \emph{pseudo-gloss} of the target rather than the raw sentence.
Given a sign video, we build token-level pseudo-gloss embeddings from a frozen language model embedding layer and temporal visual embeddings from a video encoder, then estimate a soft alignment between visual positions and pseudo-gloss tokens by solving an entropy-regularized optimal transport problem~\cite{cuturi2013sinkhorn} with a learnable null token.
The null token absorbs transitional gestures and co-articulation that correspond to no explicit word, stabilizing alignment on continuous signing.
Using the resulting transport plan, we reorder the visual features into a sequence that follows the target token order and feed it to the language model decoder.
At inference the target token order is unknown, so the visual features are not reordered; the decoder generates pseudo-gloss directly from the signing-order features, and a lightweight text-only recovery model restores spoken word order and the omitted function words.

VTaMo is trained end to end with three complementary alignment objectives beyond the standard translation loss.
First, a \emph{local} optimal transport loss encourages fine-grained correspondence between temporal visual segments and text tokens while allowing unaligned frames to map to the null token.
Second, a \emph{global} loss calibrates the sentence-level geometry of the visual and textual embedding spaces through a learnable orthogonal transformation, supported by a memory queue that diversifies the sentence pairs used for alignment.
Third, a \emph{position-aligned contrastive} loss~\cite{oord2018infonce} sharpens token-level discriminability between reordered visual features and their text token embeddings without altering the language model embedding space.
Together, these losses make the reordered visual features semantically coherent, simplifying decoder learning and improving robustness across datasets.

We evaluate on four widely used benchmarks spanning three languages and diverse data regimes: Phoenix-2014T (German), CSL-Daily (Chinese), How2Sign and OpenASL (English).
Across all benchmarks, VTaMo achieves state-of-the-art performance under the gloss-free setting and yields consistent gains over strong baselines that decode without explicit alignment.
Ablation studies verify that each alignment component contributes complementary improvements, and qualitative analysis of the learned transport plans reveals interpretable correspondences between signing segments and target tokens.

\section{Related Work}
\label{sec:related}

\subsection{Gloss-free SLT, Alignment, and Large-Scale Pretraining}
Gloss-free sign language translation (SLT) translates continuous sign videos into spoken-language sentences without gloss supervision, but it typically underperforms gloss-based pipelines due to the absence of explicit intermediate structure \cite{camgoz2020sign, zhou2021improving, zhou2021spatial, yin2021simul, chen2022simple, chen2022twostream, zhang2023sltunet, jing2024vk}. Prior work improves gloss-free SLT by strengthening temporal modeling \cite{li2020tspnet}, introducing cross-modal alignment objectives \cite{zhao2021conditional, lin2023gloss, fu2023token, jiao2024vap}, incorporating large language models \cite{wong2024sign2gpt, gong2024llms, chen2024factorized}, scaling training data \cite{uthus2024youtube, rust2024towards}, and large-scale sign pretraining such as ShuBERT~\cite{gueuwou2025shubert} and Uni-Sign~\cite{li2025unisign}; SpaMo~\cite{hwang2025efficient} further emphasizes the non-trivial sign--text correspondence via contrastive objectives. These approaches—including VAP~\cite{jiao2024vap}, which applies text-derived constraints only as visual pre-training—keep sentence-internal correspondence implicit and rely on coarse decoder attention to resolve word order. Our method instead estimates token-level alignment and reorders visual features before decoding via Sinkhorn OT with a learnable null assignment and complementary local and global objectives.

\subsection{Cross-Modal and Video--Text Alignment}
Our objectives build on general cross-modal alignment tools but adapt them to the structure of sign language. Entropy-regularized optimal transport with the Sinkhorn algorithm~\cite{cuturi2013sinkhorn} is a standard device for matching features across modalities, e.g., aligning visual features with text prompts in vision--language models~\cite{chen2023plot}; learnable orthogonal transformations relate two embedding spaces without distorting their geometry, as in cross-lingual word-embedding alignment~\cite{lample2018word}; and contrastive video--text frameworks such as CLIP4Clip~\cite{luo2022clip4clip}, VideoCLIP~\cite{xu2021videoclip}, and X-CLIP~\cite{ma2022xclip} learn joint embeddings for retrieval. Closer in spirit, procedure-learning and step-grounding methods align video frames with an ordered sequence of procedure steps, e.g., grounding instructional-article steps via narrations~\cite{mavroudi2023learning} and optimal-transport-guided procedure learning~\cite{chowdhury2024opel}. These methods, however, target global or clip-level correspondence, or assume a largely fixed, monotonic step order. Sign language translation differs: the mapping is \emph{non-monotonic} at the lexical level and \emph{partial}, since many frames are transitional and words have no manual counterpart. VTaMo addresses both by coupling OT with a null assignment, sentence-level orthogonal calibration, and position-aligned contrastive learning to yield a reordered visual sequence.


\section{Method}
\label{sec:method}

\begin{figure*}[t]
    \centering
    \includegraphics[width=\textwidth]{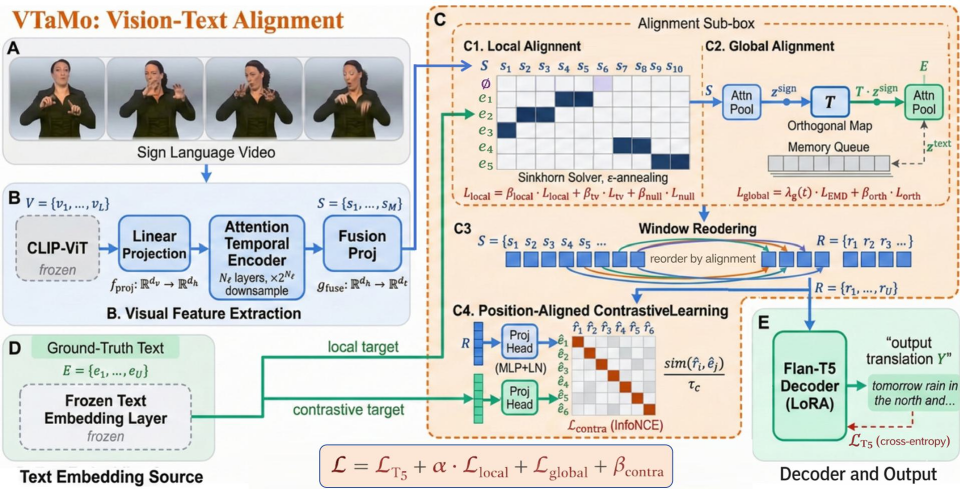}
    \caption{\textbf{VTaMo pipeline.} A sign-language video is encoded into visual tokens using a frozen CLIP-ViT backbone followed by a lightweight temporal encoder and fusion projection (\textbf{A,B}). Given the ground-truth text embeddings (\textbf{D}), VTaMo performs \textbf{local alignment} with an entropy-regularized OT solver (Sinkhorn) to obtain soft token--frame matches, and \textbf{global alignment} by mapping sign features into the text space with an orthogonal transform and a memory queue (\textbf{C1,C2}). The resulting correspondence is used for \textbf{window reordering} (\textbf{C3}) and \textbf{position-aligned contrastive learning} (\textbf{C4}). A LoRA-adapted Flan-T5 decoder generates the translation (\textbf{E}). Video frames are from the Phoenix-2014T dataset~\cite{camgoz2018neural}.}
    \label{fig:overview}
\end{figure*}

We present \textbf{VTaMo}, a framework for gloss-free sign language translation that explicitly aligns and reorders visual features before text generation.
As illustrated in Fig.~\ref{fig:overview}, VTaMo introduces three complementary alignment objectives: (1)~\emph{local alignment} via entropy-regularized optimal transport for fine-grained frame-to-token correspondences; (2)~\emph{global alignment} via a learnable orthogonal transformation to calibrate the embedding spaces at the sentence level; and (3)~\emph{position-aligned contrastive learning} for discriminative representations.

\subsection{Problem Formulation}
\label{sec:overview}

Given a sign language video, we extract frame-level features $\mathbf{V} = \{\mathbf{v}_1, \dots, \mathbf{v}_L\}$ using a pre-trained CLIP-ViT encoder, where $\mathbf{v}_i \in \mathbb{R}^{d_v}$.
The goal is to produce the spoken language sentence corresponding to the video.

Spoken sentences contain many function words that have no direct sign-level realization, which makes a strict frame-to-word alignment ill-posed.
We therefore align against, and decode, a \emph{pseudo-gloss} sequence $\mathbf{Y} = \{y_1, \dots, y_U\}$ of $U$ tokens derived from the target sentence by a fixed offline filter.
Concretely, we apply a frozen spaCy part-of-speech tagger that retains sign-relevant content tokens (\texttt{NOUN}, \texttt{VERB}, \texttt{ADJ}, \texttt{ADV}, \texttt{NUM}, \texttt{PRON}, \texttt{PROPN}) and removes tokens that usually lack a direct sign counterpart (\texttt{DET}, \texttt{ADP}, \texttt{AUX}, \texttt{PART}, \texttt{PUNCT}, \texttt{CCONJ}); the same filtering rule is applied across all datasets using language-specific spaCy models.
Both the text tokens used for alignment and the decoder targets are pseudo-gloss, and a lightweight text-only recovery model restores fluent sentences at inference.

The visual features are projected to $\mathbb{R}^{d_h}$ via a linear layer, temporally downsampled by an attention-based encoder to produce $\mathbf{H} = \{\mathbf{h}_1, \dots, \mathbf{h}_M\}$ with $M \ll L$, and mapped to the language model dimension $d_t$ by a fusion projector.
A LoRA-adapted Flan-T5 model then autoregressively generates the pseudo-gloss sequence.
The central challenge is that sign language does not follow the word order of spoken language, and the temporal granularity of visual frames far exceeds that of linguistic tokens.

\subsection{Attention-Based Temporal Encoding}
\label{sec:temporal_encoder}
We use an attention-based temporal convolution module with $N_\ell$ cascaded layers, each performing $2\times$ downsampling via learnable attention-weighted aggregation over local windows of size $W$.
For each output position $i$, a local window $\mathcal{W}_i$ centered at position $2i$ is extracted, and a query is formed from the mean-pooled local context plus a learnable positional embedding $\mathbf{p}$:
\begin{equation}
  \mathbf{q}_i = \frac{1}{|\mathcal{W}_i|}\sum_{j \in \mathcal{W}_i}\mathbf{x}_j + \mathbf{p}.
  \label{eq:temporal_query}
\end{equation}
Multi-head attention with relative position bias then aggregates the local context:
\begin{equation}
  \alpha_{i,j} = \mathrm{softmax}\!\left(\frac{\mathbf{q}_i^\top \mathbf{k}_j}{\sqrt{d_h / N_\text{head}}} + r_{j-c_i}\right),
  \qquad
  \mathbf{o}_i = \sum_{j \in \mathcal{W}_i}\alpha_{i,j}\,\mathbf{v}_j,
  \label{eq:temporal_attn}
\end{equation}
where $c_i$ is the window center, $r_{j-c_i}$ is a learnable relative position bias, and $N_\text{head}$ is the number of attention heads.
This ensures gradient flow to all frames, yielding more expressive temporal representations than max-pooling.

\subsection{Local Alignment via Optimal Transport}
\label{sec:local_alignment}

Local alignment is the core component of VTaMo, establishing fine-grained correspondences between temporal visual segments and text tokens, formulated as an entropy-regularized optimal transport problem.
Let $\mathbf{S} = \{\mathbf{s}_1, \dots, \mathbf{s}_M\} \in \mathbb{R}^{M \times d_t}$ denote the visual features after fusion projection, and $\mathbf{E} = \{\mathbf{e}_1, \dots, \mathbf{e}_U\} \in \mathbb{R}^{U \times d_t}$ the pseudo-gloss token embeddings from the frozen encoder embedding layer.
Since $\mathbf{S}$ is optimized against the fixed text embeddings $\mathbf{E}$ under the alignment and translation losses, their cosine similarity reflects sign-level semantic correspondence, making it a meaningful transport cost.
We introduce a single learnable null token $\mathbf{e}_\varnothing \in \mathbb{R}^{d_t}$, prepended at position $0$ of the text sequence, forming an augmented sequence $\tilde{\mathbf{E}} \in \mathbb{R}^{K \times d_t}$ with $K = U + 1$.
The assignment is deliberately not one-to-one: frames that carry no explicit sign-language semantics---such as background, transition, or reset frames---are all allowed to align to this same null token, so the model is never forced to map every frame onto a content token.
The pairwise cost matrix uses cosine distance, with a learnable bias $b_\varnothing$ controlling null affinity:
\begin{equation}
  C_{m,k} =
  \begin{cases}
    1 - \bar{\mathbf{s}}_m^\top \bar{\mathbf{e}}_k - b_\varnothing, & \text{if } k = 0 \text{ (null)}, \\
    1 - \bar{\mathbf{s}}_m^\top \bar{\mathbf{e}}_k, & \text{otherwise},
  \end{cases}
  \label{eq:cost_matrix}
\end{equation}
where $\bar{\mathbf{s}}_m$ and $\bar{\mathbf{e}}_k$ are $\ell_2$-normalized vectors.
We solve the entropy-regularized OT problem via the Sinkhorn algorithm in log-domain:
\begin{equation}
  \mathbf{A}^* = \arg\min_{\mathbf{A} \in \Pi(\mathbf{a}, \mathbf{b})} \; \langle \mathbf{A}, \mathbf{C} \rangle - \varepsilon\, H(\mathbf{A}),
  \label{eq:ot_problem}
\end{equation}
where $\Pi(\mathbf{a}, \mathbf{b})$ is the set of transport plans with uniform marginals over valid frame and token positions, $H(\mathbf{A}) = -\sum_{m,k} A_{m,k}\log A_{m,k}$ is the entropy term, and $\varepsilon$ controls the sharpness of assignments.
For numerical stability we iterate the dual potentials in the log domain for $N_\text{iter}$ steps,
\begin{align}
  \log u_m &\leftarrow \log a_m - \log\textstyle\sum_k \exp\!\left(-C_{m,k}/\varepsilon + \log v_k\right), \nonumber \\
  \log v_k &\leftarrow \log b_k - \log\textstyle\sum_m \exp\!\left(-C_{m,k}/\varepsilon + \log u_m\right),
  \label{eq:sinkhorn_dual}
\end{align}
and recover the plan as $A_{m,k} = \exp\!\left(\log u_m - C_{m,k}/\varepsilon + \log v_k\right)$.
We adopt a multi-phase annealing schedule that progressively decreases $\varepsilon$ from a high initial value to encourage exploration before converging to sharp, discrete assignments.

The local alignment loss combines three terms.
The first is the transport cost $\mathcal{L}_\text{trans} = \frac{1}{B}\sum_i \sum_{m,k} A_{m,k}^{(i)} C_{m,k}^{(i)}$.
The second is a temporal variation regularization that penalizes adjacent frames switching tokens, computed on the row-normalized plan $\hat{\mathbf{A}}$ restricted to real text tokens:
\begin{equation}
  \mathcal{L}_\text{tv} = \frac{1}{(M-1)(K-1)}\sum_{m=1}^{M-1}\sum_{k=1}^{K-1}\big|\hat{A}_{m+1,k} - \hat{A}_{m,k}\big|.
  \label{eq:tv_loss}
\end{equation}
The third is a null cost regularization that keeps the mean cost-based null affinity $\bar{p}_\varnothing$---computed via a row-wise softmax on the cost matrix independently of the transport plan---below a target ratio $\rho_\text{target}$, preventing the learnable bias $b_\varnothing$ from making the null column uniformly cheaper than all real tokens:
\begin{equation}
  \mathcal{L}_\text{null} = \big[\bar{p}_\varnothing - \rho_\text{target}\big]_{+}^{2},
  \qquad
  \bar{p}_\varnothing = \frac{1}{M}\sum_{m=1}^{M} \frac{\exp(-C_{m,0}/\tau)}{\sum_{k=0}^{K-1}\exp(-C_{m,k}/\tau)}.
  \label{eq:null_loss}
\end{equation}
The complete local objective is their weighted sum:
\begin{equation}
  \mathcal{L}_\text{local} = \beta_\text{local}\, \mathcal{L}_\text{trans} + \beta_\text{tv}\, \mathcal{L}_\text{tv} + \beta_\text{null}\, \mathcal{L}_\text{null}.
  \label{eq:total_local_loss}
\end{equation}

\subsection{Global Alignment via Orthogonal Transformation}
\label{sec:global_alignment}

Because the visual and textual encoders are pre-trained on different modalities, paired sign and text sentence embeddings can suffer a coordinate (orientation) mismatch even when they are semantically equivalent.
We therefore apply a learnable \emph{orthogonal transformation}---a rotation of the feature space that preserves vector lengths and angular relationships---to the visual sentence embeddings before comparing the two modalities at the sentence level.
We constrain the transformation to be orthogonal precisely because the goal is to correct orientation without reshaping the space: this preserves norms and angular structure for the cosine-based transport cost and avoids the spurious low-cost matches that an unconstrained projection could introduce.

We aggregate frame-level features into one sentence-level vector per video and per sentence via attention pooling, using a mean query $\mathbf{q}$ over valid positions and the normalized weighted sum
\begin{equation}
  \mathbf{z} = \frac{\sum_l \alpha_l \mathbf{x}_l}{\big\|\sum_l \alpha_l \mathbf{x}_l\big\|_2},
  \qquad
  \alpha_l = \mathrm{softmax}_l\!\big(\mathbf{x}_l^\top \mathbf{q}/\sqrt{d_t}\big),
  \label{eq:attn_pool}
\end{equation}
yielding normalized sentence-level vectors $\mathbf{z}^\text{sign}, \mathbf{z}^\text{text} \in \mathbb{R}^{d_t}$.
A FIFO memory queue stores recent sentence-level vector pairs so that the sentence-level transport is computed over a sufficiently diverse batch rather than a few in-batch pairs.
A learnable matrix $\mathbf{T} \in \mathbb{R}^{d_t \times d_t}$, initialized near identity, transforms the visual sentence vectors:
\begin{equation}
  \tilde{\mathbf{z}}_n^\text{sign} = \frac{\mathbf{z}_n^\text{sign}\, \mathbf{T}}{\|\mathbf{z}_n^\text{sign}\, \mathbf{T}\|_2}.
  \label{eq:global_transform}
\end{equation}
We then form the pairwise cosine cost $C^g_{n,n'} = 1 - (\tilde{\mathbf{z}}_n^\text{sign})^\top \mathbf{z}_{n'}^\text{text}$ between the transformed visual vectors and the textual vectors and solve a Sinkhorn OT problem with uniform marginals to obtain a sentence-level transport plan $\mathbf{P}$; the resulting Earth Mover's Distance loss is
\begin{equation}
  \mathcal{L}_\text{EMD} = \sum_{n,n'} P_{n,n'}\, (1 - (\tilde{\mathbf{z}}_n^\text{sign})^\top \mathbf{z}_{n'}^\text{text}).
  \label{eq:emd_loss}
\end{equation}
An orthogonality constraint $\mathcal{L}_\text{orth} = \|\mathbf{T}^\top \mathbf{T} - \mathbf{I}\|_F^2$ keeps $\mathbf{T}$ close to a pure rotation and preserves pairwise distances.
The global loss is activated only after local alignment stabilizes, using a scheduled ramp-up:
\begin{equation}
  \mathcal{L}_\text{global} = \lambda_g(t) \cdot \mathcal{L}_\text{EMD} + \beta_\text{orth}\, \mathcal{L}_\text{orth}.
  \label{eq:total_global_loss}
\end{equation}

\subsection{Window-Based Feature Reordering}
\label{sec:reordering}

Once local alignment has produced a reliable transport plan $\mathbf{A}$, we reorder the visual features to match the target token order.
This is necessary because the decoder's autoregressive cross-entropy loss expects the input to follow the target text order; otherwise the visual features stay in temporal order and the cross-entropy gradient counteracts the alignment objectives.

Concretely, we partition the $M$ visual frames into $N_w = U + 2$ overlapping temporal windows, where the two extra windows absorb boundary effects; the $w$-th window spans $[s_w, e_w)$ with $s_w = \lfloor w\,(M-1)/N_w + 0.5\rfloor$ and $e_w = \lfloor (w+1)(M-1)/N_w + 0.5\rfloor + 1$.
For each window we accumulate the alignment mass $\boldsymbol{\pi}_w = \sum_{m=s_w}^{e_w-1}\mathbf{A}_{m,:} \in \mathbb{R}^{K}$ and assign it to the real text token with the largest aggregated mass, $\phi(w) = \arg\max_{u \in \{1,\dots,U\}} \pi_{w,u}$; windows dominated by the null token are excluded.
Collecting the windows assigned to each token $u$ into $\Omega_u = \{w \mid \phi(w) = u\}$ and traversing the token order left to right yields the reordered sequence
\begin{equation}
  \mathbf{R} = \mathrm{Concat}\big(\{\mathbf{S}_{s_w:e_w-1} \mid w \in \Omega_1^{\uparrow}\}, \dots, \{\mathbf{S}_{s_w:e_w-1} \mid w \in \Omega_U^{\uparrow}\}\big) \in \mathbb{R}^{U' \times d_t},
  \label{eq:window_reorder}
\end{equation}
where $\Omega_u^{\uparrow}$ orders the windows in $\Omega_u$ by their original temporal index.
This preserves local temporal continuity, and the reordered sequence is fed to the decoder.
For the contrastive objective, frames assigned to each token $u$ are mean-pooled to produce a single token-level representation $\mathbf{r}_u$, yielding $\mathbf{R} = \{\mathbf{r}_1, \dots, \mathbf{r}_U\}$ aligned one-to-one with $\mathbf{E}$.

\subsection{Position-Aligned Contrastive Learning}
\label{sec:contrastive}

Because reordering is used only during training, the model must also function at inference when the target order is unknown.
We therefore impose a token-level contrastive objective that binds each reordered visual feature to its text token embedding, so that even when tokens arrive out of order each is paired with its correct counterpart and the decoder can recover its lexical item; restoring fluent word order is then left to the recovery model.

We apply learnable projection heads to both modalities:
\begin{equation}
  \hat{\mathbf{r}}_u = \text{LN}\!\big(\mathbf{W}_2^\text{v}\, \sigma(\mathbf{W}_1^\text{v}\, \mathbf{r}_u)\big), \quad
  \hat{\mathbf{e}}_u = \text{LN}\!\big(\mathbf{W}_2^\text{t}\, \sigma(\mathbf{W}_1^\text{t}\, \mathbf{e}_u)\big),
\end{equation}
where $\sigma$ is GELU activation and $\text{LN}$ is layer normalization.
After $\ell_2$-normalization, the InfoNCE loss is:
\begin{equation}
  \mathcal{L}_\text{contra} = -\frac{1}{|\mathcal{V}|}\sum_{i \in \mathcal{V}} \log \frac{\exp\!\big(\text{sim}(\hat{\mathbf{r}}_i, \hat{\mathbf{e}}_i) / \tau_c\big)}{\displaystyle\sum_{j \in \mathcal{V}} \exp\!\big(\text{sim}(\hat{\mathbf{r}}_i, \hat{\mathbf{e}}_j) / \tau_c\big)},
  \label{eq:contrastive_loss}
\end{equation}
where $\mathcal{V}$ denotes valid token positions and $\tau_c$ is a temperature hyperparameter.
Gradients flow through the visual branch but are blocked from the text branch to preserve the pre-trained language model representations.

\paragraph{Inference and pseudo-gloss recovery.}
At test time the target order is unknown, so no reordering is applied: the decoder reads the visual tokens in their original temporal order and emits a pseudo-gloss sequence in visual (signing) order rather than spoken-language order.
The token-level grounding above makes this feasible, since each visual token is decoded into its correct lexical item even when the tokens arrive out of order.
To recover a fluent translation, we pass this video-order pseudo-gloss through a lightweight recovery model that restores spoken word order and re-inserts the function words dropped by the pseudo-gloss filter (Sec.~\ref{sec:overview}).
Crucially, this recovery model is trained purely on text and never sees sign videos: we take plain sentences, apply the same pseudo-gloss extractor, permute the content words, and train the model to reconstruct the original sentence from this shuffled input.

The overall training objective combines the translation loss with the three alignment losses:
\begin{equation}
  \mathcal{L} = \mathcal{L}_\text{T5} + \lambda_\text{local}\, \mathcal{L}_\text{local} + \lambda_g(t)\, \mathcal{L}_\text{EMD} + \beta_\text{orth}\, \mathcal{L}_\text{orth} + \beta_\text{contra}\, \mathcal{L}_\text{contra},
  \label{eq:total_loss}
\end{equation}
where $\mathcal{L}_\text{T5}$ is the cross-entropy loss for autoregressive generation.
Training proceeds in two phases: in the first phase, only alignment losses are active while the language model is frozen, allowing the visual encoder to establish meaningful correspondences; in the second phase, the full objective is optimized jointly with the LoRA-adapted language model.


\section{Experiments}
\label{sec:experiments}

\subsection{Datasets and Metrics}
\label{sec:datasets}

We evaluate our method on four publicly available sign language translation benchmarks that span different languages, vocabulary sizes, and recording conditions: Phoenix-2014T~\cite{camgoz2018neural} (German), CSL-Daily~\cite{zhou2021improving} (Chinese), How2Sign~\cite{duarte2021how2sign} and OpenASL~\cite{shi2022open} (English).
We report BLEU-$n$ scores~\cite{papineni2002bleu} for $n \in \{1, 2, 3, 4\}$ and ROUGE-L F1~\cite{lin2004rouge} as the primary evaluation metrics.
For English SLT datasets (How2Sign and OpenASL), we additionally report BLEURT~\cite{sellam2020bleurt} scores following established protocol.
For every benchmark, each model trains only on its official split without additional sign-language data or task-specific pre-training.

\subsection{Implementation Details}
\label{sec:impl}

\paragraph{Visual features.}
We extract frame-level spatial features using a pre-trained CLIP-ViT-Large model~\cite{radford2021clip} with S$^2$-Wrapper~\cite{shi2024when} for multi-scale feature extraction at scales 1 and 2, yielding 2048-dimensional feature vectors per frame.

\paragraph{Language model.}
We employ Flan-T5-XL~\cite{chung2024scaling} as the language model backbone and adapt it using LoRA~\cite{hu2022lora} with rank $r = 16$, scaling factor $\alpha = 32$, and dropout rate 0.1.
The LoRA adapters are applied to the query and value projection matrices across all transformer layers.

\paragraph{Architecture.}
The visual features are projected from $d_v = 2048$ to an intermediate dimension $d_h = 768$ through a linear layer.
The attention-based temporal encoder consists of $N_\ell = 2$ cascaded layers with 4 attention heads and a window size of 8, yielding a $4\times$ temporal downsampling.
The fusion projector maps the features from $d_h$ to the T5 embedding dimension $d_t$.
The contrastive projection heads are two-layer MLPs with GELU activation and layer normalization, projecting to 768 dimensions.

\paragraph{Training.}
We train all models using AdamW~\cite{loshchilov2019decoupled} with a peak learning rate of $6 \times 10^{-4}$, cosine learning rate schedule, and linear warm-up over the first 2,000 steps.
The batch size is 6 with gradient accumulation over 2 steps, and gradient clipping is applied with a maximum norm of 1.0.
All experiments were conducted on a single NVIDIA A100 GPU with mixed-precision (bf16) training.

\paragraph{Alignment hyperparameters.}
For the local alignment, the Sinkhorn algorithm runs for $N_\text{iter} = 10$ iterations.
The epsilon annealing schedule proceeds from $\varepsilon_\text{high} = 0.12$ to $\varepsilon_\text{mid} = 0.10$ over the first 10 epochs and further to $\varepsilon_\text{low} = 0.03$ over the following 80 epochs.
The null ratio target is $\rho_\text{target} = 0.2$, and the loss weights are $\beta_\text{local} = 2.0$, $\beta_\text{tv} = 0.1$, $\beta_\text{null} = 0.1$, with the overall local-alignment weight $\lambda_\text{local} = 1.0$.
For the global alignment, the EMD solver uses $\varepsilon_g = 0.05$ with 20 Sinkhorn iterations, the memory queue capacity is 256, and the weight ramps from 0 to $\lambda_g^\text{max} = 0.1$ over 4,000 steps after warm-up.
The orthogonality penalty coefficient is $\beta_\text{orth} = 0.05$.
The contrastive learning temperature is $\tau_c = 0.1$, and the contrastive loss weight is $\beta_\text{contra} = 1.0$.

\paragraph{Training cost.}
On a single A100, end-to-end training takes 14\,h on Phoenix-2014T, 32\,h on CSL-Daily, and 50\,h on How2Sign.
The optimal transport solvers add negligible overhead: the local Sinkhorn step takes 54\,ms ($<$3.2\% of the per-step time), and the global OT over the $256\times256$ memory queue takes $<$38\,ms.

\subsection{Comparison with State-of-the-Art}

\textbf{Results on Phoenix-2014T and CSL-Daily.} As shown in Table~\ref{tab:phoenix_csl}, most prior methods rely on visual-encoder fine-tuning or gloss supervision, yet VTaMo reaches state-of-the-art on both benchmarks without either. On Phoenix-2014T it attains 28.86 BLEU-4, surpassing the strongest gloss-free method SpaMo~\cite{hwang2025efficient} by 4.54 points and coming within 0.09 of the best gloss-based result (TS-SLT~\cite{chen2022twostream}, 28.95). On CSL-Daily it reaches 27.16 BLEU-4, a 6.61-point gain over SpaMo and 0.80 over Uni-Sign, showing that explicit multi-granularity alignment especially helps on topically diverse data.

\begin{table*}[t]
\centering
\caption{Performance comparison on the Phoenix-2014T~\cite{camgoz2018neural} and CSL-Daily datasets~\cite{zhou2021improving}; all numbers are reported on the official test sets. ``Vis.Ft.'' denotes visual fine-tuning on sign language datasets. Bold marks the best result.}
\label{tab:phoenix_csl}
\resizebox{\textwidth}{!}{
\begin{tabular}{lllccccccccccc}
\toprule
\multirow{2}{*}{Setting} & \multirow{2}{*}{Methods} & \multirow{2}{*}{Modality} & \multirow{2}{*}{Vis.Ft.} & \multicolumn{5}{c}{Phoenix-2014T} & \multicolumn{5}{c}{CSL-Daily} \\
\cmidrule(lr){5-9} \cmidrule(lr){10-14}
 & & & & B1 & B2 & B3 & B4 & RG & B1 & B2 & B3 & B4 & RG \\
\midrule
\multirow{4}{*}{Gloss-based}
 & SLRT~\cite{camgoz2020sign} & RGB & \cmark & 46.61 & 33.73 & 26.19 & 21.32 & 49.54 & 37.38 & 24.36 & 16.55 & 11.79 & 36.74 \\
 & MMTLB~\cite{chen2022simple} & RGB & \xmark & 53.97 & 41.75 & 33.84 & 28.39 & 52.65 & 53.31 & 40.41 & 30.87 & 23.92 & 53.25 \\
 & TS-SLT~\cite{chen2022twostream} & RGB & \cmark & 54.90 & 42.43 & 34.46 & \textbf{28.95} & 53.48 & 55.44 & 42.59 & 32.87 & 25.79 & 55.72 \\
 & SLTUNET~\cite{zhang2023sltunet} & RGB & \cmark & 52.92 & 41.76 & 33.99 & 28.47 & 52.11 & 54.98 & 41.44 & 31.84 & 25.01 & 54.08 \\
\midrule
\multirow{5}{*}{Weakly Gloss-free}
 & TSPNet~\cite{li2020tspnet} & RGB & \cmark & 36.10 & 23.12 & 16.88 & 13.41 & 34.96 & 17.09 & 8.98 & 5.07 & 2.97 & 18.38 \\
 & GASLT~\cite{yin2023gloss} & RGB & \cmark & 39.07 & 26.74 & 21.86 & 15.74 & 39.86 & 19.90 & 9.94 & 5.98 & 4.07 & 20.35 \\
 & CSGCR~\cite{zhao2021conditional} & RGB & \xmark & 36.71 & 25.40 & 18.86 & 15.18 & 38.85 & --- & --- & --- & --- & --- \\
 & GFSLT-VLP~\cite{zhou2023gloss} & RGB & \cmark & 43.71 & 33.18 & 26.11 & 21.44 & 42.49 & 39.37 & 24.93 & 16.26 & 11.00 & 36.44 \\
 & FLa-LLM~\cite{chen2024factorized} & RGB & \xmark & 46.29 & 35.33 & 28.03 & 23.09 & 45.27 & 37.15 & 25.12 & 18.38 & 14.20 & 37.25 \\
\midrule
\multirow{6}{*}{Gloss-free}
 & Sign2GPT~\cite{wong2024sign2gpt} & RGB & \cmark & 49.54 & 35.96 & 28.83 & 22.52 & 48.90 & 41.75 & 28.73 & 20.60 & 15.40 & 42.36 \\
 & SignLLM~\cite{gong2024llms} & RGB & \cmark & 45.21 & 34.78 & 28.05 & 23.40 & 44.49 & 39.55 & 28.13 & 20.07 & 15.75 & 39.91 \\
 & C$^2$RL~\cite{chen2024c2rl} & RGB & \cmark & 52.81 & 40.20 & 32.20 & 26.75 & 50.96 & 49.32 & 36.28 & 27.54 & 21.61 & 48.21 \\
 & Uni-Sign~\cite{li2025unisign} & Pose+RGB & \cmark & --- & --- & --- & --- & --- & 55.08 & 42.14 & 32.98 & 26.36 & 56.51 \\
 & SpaMo~\cite{hwang2025efficient} & RGB & \xmark & 49.80 & 37.32 & 29.50 & 24.32 & 46.57 & 48.90 & 36.90 & 26.78 & 20.55 & 47.46 \\
 & VTaMo (Ours) & RGB & \xmark & \textbf{61.32} & \textbf{46.27} & \textbf{35.15} & 28.86 & \textbf{60.24} & \textbf{57.64} & \textbf{44.12} & \textbf{33.78} & \textbf{27.16} & \textbf{57.28} \\
\bottomrule
\end{tabular}
}
\end{table*}

\textbf{Results on How2Sign.} How2Sign is more challenging, with open-domain topics and a much larger vocabulary. As shown in Table~\ref{tab:how2sign}, VTaMo achieves the best BLEU-1, BLEU-4, and BLEURT on the official test set (BLEU-4 18.47, BLEURT 50.26) using only RGB and no large-scale pre-training---an 8.36 BLEU-4 gain over SpaMo~\cite{hwang2025efficient}---and ranks second on ROUGE-L behind SSVP-SLT~\cite{rust2024towards} (38.40 vs.\ 37.14). It also surpasses SHuBERT~\cite{gueuwou2025shubert} (16.2 / 49.9) and Uni-Sign~\cite{li2025unisign} (14.9 / 49.4), both using pose+RGB and large-scale pre-training.

\textbf{Results on OpenASL.} OpenASL is the hardest setting, with a vocabulary exceeding 10{,}000 words. As shown in Table~\ref{tab:openasl}, VTaMo attains 25.94 BLEU-4 and 62.48 BLEURT on the official test set, outperforming Uni-Sign~\cite{li2025unisign} and SHuBERT~\cite{gueuwou2025shubert} by 2.80 and 2.74 BLEU-4 despite their pose+RGB inputs and large-scale pre-training. Gains over SpaMo grow on the larger-vocabulary How2Sign and OpenASL, consistent with alignment helping more as complexity increases.

\begin{table*}[t]
\centering
\begin{minipage}[t]{0.48\textwidth}
\centering
\caption{Results on How2Sign~\cite{duarte2021how2sign}. Bold marks the best result.}
\label{tab:how2sign}
\resizebox{\linewidth}{!}{
\begin{tabular}{llcccc}
\toprule
Methods & Mod. & B1 & B4 & RG & BLEURT \\
\midrule
GloFE-VN~\cite{lin2023gloss} & Lmk & 14.9 & 2.2 & 12.6 & 31.7 \\
YT-ASL~\cite{uthus2024youtube} & Lmk & 14.96 & 1.22 & 25.70 & 29.98 \\
SSVP-SLT~\cite{rust2024towards} & RGB & 43.20 & 15.50 & \textbf{38.40} & 49.60 \\
FLa-LLM~\cite{chen2024factorized} & RGB & 29.81 & 9.66 & 27.81 & --- \\
Uni-Sign~\cite{li2025unisign} & P+R & 40.2 & 14.9 & 36.0 & 49.4 \\
SHuBERT~\cite{gueuwou2025shubert} & P+R & --- & 16.2 & --- & 49.9 \\
SpaMo~\cite{hwang2025efficient} & RGB & 33.41 & 10.11 & 30.56 & 42.23 \\
\midrule
VTaMo (Ours) & RGB & \textbf{46.67} & \textbf{18.47} & 37.14 & \textbf{50.26} \\
\bottomrule
\end{tabular}
}
\end{minipage}
\hfill
\begin{minipage}[t]{0.48\textwidth}
\centering
\caption{Results on OpenASL~\cite{shi2022open}. Bold marks the best result.}
\label{tab:openasl}
\resizebox{\linewidth}{!}{
\begin{tabular}{llcccc}
\toprule
Methods & Mod. & B1 & B4 & RG & BLEURT \\
\midrule
GloFE-VN~\cite{lin2023gloss} & Pose & 21.56 & 7.06 & 21.75 & 36.35 \\
Conv-GRU~\cite{camgoz2018neural} & RGB & 16.11 & 4.58 & 16.10 & 25.65 \\
OpenASL~\cite{shi2022open} & RGB & 20.92 & 6.72 & 21.02 & 31.09 \\
C$^2$RL~\cite{chen2024c2rl} & RGB & 31.46 & 13.21 & 31.36 & --- \\
Uni-Sign~\cite{li2025unisign} & P+R & 49.35 & 23.14 & 43.22 & 60.40 \\
SHuBERT~\cite{gueuwou2025shubert} & P+R & --- & 23.2 & --- & 60.60 \\
SpaMo~\cite{hwang2025efficient} & RGB & 46.28 & 21.25 & 41.04 & 57.82 \\
\midrule
VTaMo (Ours) & RGB & \textbf{53.58} & \textbf{25.94} & \textbf{46.85} & \textbf{62.48} \\
\bottomrule
\end{tabular}
}
\end{minipage}
\end{table*}

\subsection{Ablation Study}
\label{sec:ablation}

All ablations are conducted on the Phoenix-2014T test set.

\paragraph{Alignment components.}
Removing each alignment loss in turn (Table~\ref{tab:ablation}) shows that all three are necessary and non-substitutable.
The position-aligned contrastive loss $\mathcal{L}_\text{contra}$ matters most ($-$6.30 BLEU-4), providing token-level discriminative grounding; the local OT loss $\mathcal{L}_\text{local}$ is second ($-$5.09), as it underpins the window-based reordering that feeds ordered inputs to the decoder; and the global loss $\mathcal{L}_\text{EMD}$ ($-$2.65) adds complementary embedding-space calibration.

\paragraph{Language-model backbone.}
To test whether the gains stem from a favorable language model rather than our alignment design, we vary the backbone following SpaMo's protocol (Table~\ref{tab:llm_ablation}).
The ranking matches SpaMo---Flan-T5-XL is best, mT0-XL is close behind, and Llama-2 underperforms despite its larger size---confirming that scaling the LM alone does not help under the limited-data SLT regime.
Critically, under \emph{every} fixed backbone our method beats SpaMo by $+2.49$ to $+4.54$ BLEU-4, so the improvement comes from the alignment block.

\paragraph{Local alignment quality.}
Three transport-plan metrics (Table~\ref{tab:alignment_metrics})---peak assignment probability, normalized entropy, and segment change rate---characterize alignment sharpness.
The full model is the sharpest and most confident; removing epsilon annealing yields diffuse assignments, removing the TV loss fragments them, and removing the null token forces transitional frames onto real tokens.

\paragraph{Global alignment components.}
Table~\ref{tab:global_ablation} isolates the global module and reports the epoch of peak development BLEU-4 as an efficiency proxy.
The orthogonality constraint prevents distance distortion, scheduled activation avoids premature gradients before local plans stabilize, and the memory queue is the largest contributor to faster convergence (36 vs.\ 68 epochs without it).

\begin{table*}[t]
\centering
\begin{minipage}[t]{0.46\textwidth}
\centering
\caption{Alignment-component ablation on the Phoenix-2014T~\cite{camgoz2018neural} test set. Each row removes one loss from the full model.}
\label{tab:ablation}
\begin{tabular}{l c c c}
\toprule
Configuration & B1 & B4 & RG \\
\midrule
Full model & \textbf{61.32} & \textbf{28.86} & \textbf{60.24} \\
\midrule
w/o $\mathcal{L}_\text{local}$ & 52.62 & 23.77 & 51.29 \\
w/o $\mathcal{L}_\text{EMD}$ & 57.46 & 26.21 & 57.92 \\
w/o $\mathcal{L}_\text{contra}$ & 51.23 & 22.56 & 50.14 \\
\bottomrule
\end{tabular}
\end{minipage}
\hfill
\begin{minipage}[t]{0.5\textwidth}
\centering
\small
\setlength{\tabcolsep}{4pt}
\caption{Language-model backbone ablation on Phoenix-2014T~\cite{camgoz2018neural} (BLEU-4), following SpaMo~\cite{hwang2025efficient}'s protocol. $\Delta$: our gain over SpaMo at the same backbone.}
\label{tab:llm_ablation}
\resizebox{\linewidth}{!}{
\begin{tabular}{l c c c c}
\toprule
Backbone & \#Param & SpaMo & Ours & $\Delta$ \\
\midrule
Flan-T5-large~\cite{chung2024scaling} & 0.78B & 19.25 & 22.07 & $+2.82$ \\
mBART-50~\cite{tang2020multilingual} & 0.61B & 10.94 & 13.43 & $+2.49$ \\
mT0-XL~\cite{muennighoff2023crosslingual} & 3.5B & 24.23 & 27.90 & $+3.67$ \\
Llama-2~\cite{touvron2023llama} & 7B & 13.86 & 16.82 & $+2.96$ \\
Flan-T5-XL~\cite{chung2024scaling} & 3.0B & 24.32 & \textbf{28.86} & $+4.54$ \\
\bottomrule
\end{tabular}
}
\end{minipage}
\end{table*}

\begin{table*}[t]
\centering
\begin{minipage}[t]{0.52\textwidth}
\centering
\caption{Alignment quality metrics on the Phoenix-2014T~\cite{camgoz2018neural} development set.}
\label{tab:alignment_metrics}
\begin{tabular}{l c c c}
\toprule
Configuration & Peak $\uparrow$ & Entropy $\downarrow$ & Change $\downarrow$ \\
\midrule
w/o annealing & 0.48 & 0.55 & 0.32 \\
w/o TV loss & 0.72 & 0.67 & 0.54 \\
w/o null token & 0.61 & 0.46 & 0.33 \\
Full model & \textbf{0.76} & \textbf{0.18} & \textbf{0.12} \\
\bottomrule
\end{tabular}
\end{minipage}
\hfill
\begin{minipage}[t]{0.44\textwidth}
\centering
\caption{Ablation of global alignment components on Phoenix-2014T~\cite{camgoz2018neural}. Ep.~$\downarrow$: best epoch.}
\label{tab:global_ablation}
\begin{tabular}{l c c}
\toprule
Configuration & B4 & Ep. $\downarrow$ \\
\midrule
Full model & \textbf{28.86} & 36 \\
w/o orthogonality & 28.13 & 52 \\
w/o mem.\ queue & 27.35 & 68 \\
w/o sched.\ act. & 26.42 & 43 \\
\bottomrule
\end{tabular}
\end{minipage}
\end{table*}

\subsection{Noise Sensitivity Analysis}
\label{sec:noise_sensitivity}

\begin{figure}[t]
\centering
\includegraphics[width=\linewidth]{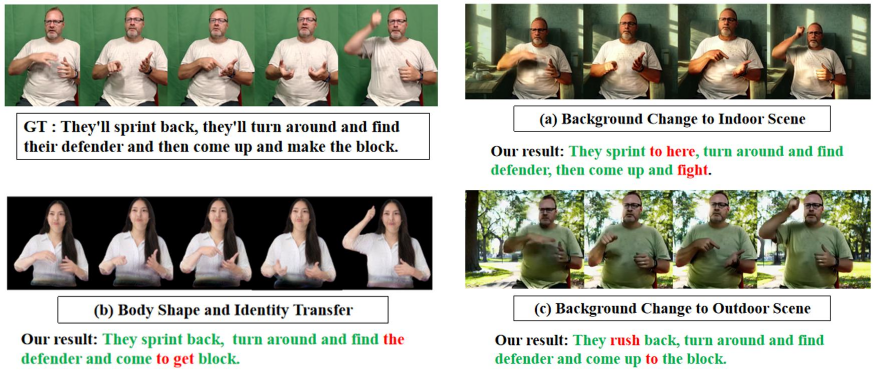}
\caption{Representative augmented samples used in the robustness analysis on How2Sign~\cite{duarte2021how2sign}. (a) and (c) are background-replacement conditions generated via text-prompt editing; (b) is a full body shape and identity transfer to a synthetic signer generated via text prompt. Hand and body kinematics are preserved via ControlNet pose conditioning. The synthetic signer in (b) does not represent any real individual. All modifications are used solely for research evaluation.}
\label{fig:four_aug}
\end{figure}

Since VTaMo operates on RGB input without visual-encoder fine-tuning, we assess its robustness to appearance shifts. Using Stable Diffusion~\cite{rombach2022ldm} with ControlNet~\cite{zhang2023controlnet} for pose-conditioned synthesis, we build four controlled conditions on the entire How2Sign test set while preserving each signer's hand and body kinematics (Fig.~\ref{fig:four_aug}): two replace the background (BG-1, outdoor park; BG-2, indoor kitchen) and two alter signer appearance (AP-1, full body shape and identity transfer to a synthetic signer; AP-2, clothing). All appearance modifications are entirely synthetic---the transferred identity in AP-1 does not represent any real individual---and are used solely for robustness evaluation. Table~\ref{tab:noise_sensitivity} reports single-run performance across all conditions: background replacement causes only marginal degradation ($\leq$0.42 BLEU-4) and appearance edits even smaller drops ($\leq$0.33), showing that VTaMo attends to gesture semantics rather than scene context or signer identity. Additional examples are in the supplementary material.

\begin{table}[t]
\centering
\caption{Noise sensitivity analysis on the entire How2Sign~\cite{duarte2021how2sign} test set. BG: background replacement, AP: appearance modification. All modifications generated with Stable Diffusion and ControlNet pose conditioning.}
\label{tab:noise_sensitivity}
\begin{tabular}{l c c c c c}
\toprule
Condition & B1 & B2 & B3 & B4 & RG \\
\midrule
Original & 46.67 & 32.47 & 24.11 & 18.47 & 37.14 \\
BG-1 (outdoor) & 46.24 & 32.12 & 23.98 & 18.26 & 37.02 \\
BG-2 (kitchen) & 45.89 & 31.48 & 23.62 & 18.05 & 36.67 \\
AP-1 (identity) & 46.15 & 32.07 & 23.91 & 18.14 & 36.92 \\
AP-2 (clothing) & 46.52 & 32.44 & 23.86 & 18.42 & 37.08 \\
\bottomrule
\end{tabular}
\end{table}

\subsection{Qualitative Analysis}
\label{sec:qualitative}

\begin{figure*}[t]
    \centering
    \includegraphics[width=\textwidth]{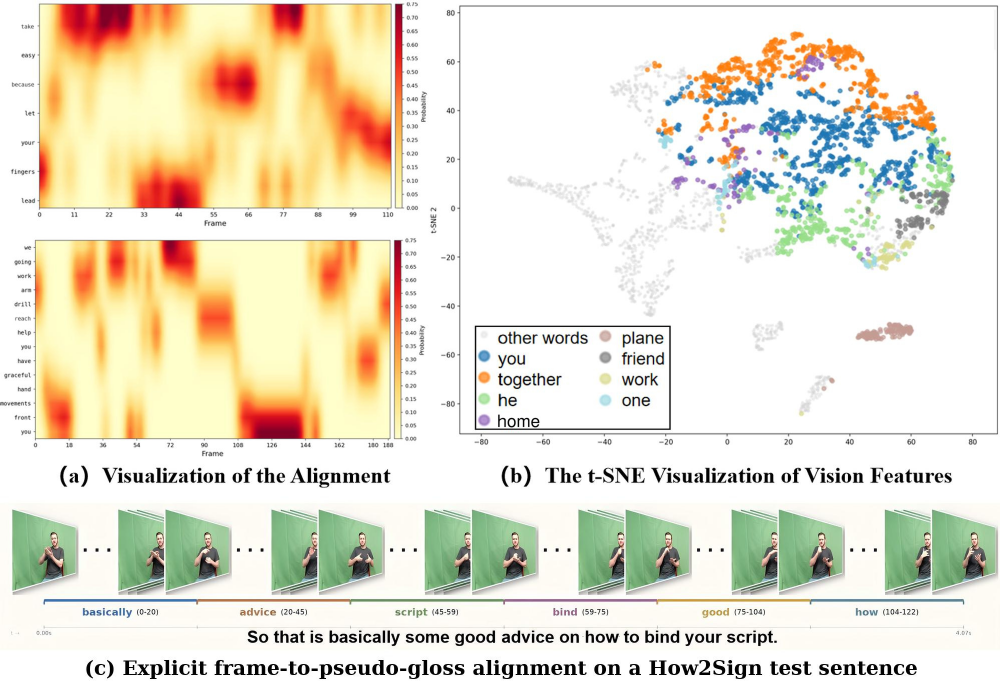}
    \caption{Qualitative analysis on How2Sign~\cite{duarte2021how2sign} test samples. (a) OT transport plans from VTaMo's Sinkhorn solver: rows are visual frames and columns are pseudo-gloss tokens, with darker cells indicating higher transport mass; the block-diagonal structure reflects temporally coherent frame-to-token correspondences while the null token absorbs transitional frames. (b) t-SNE of visual features, colored by semantic category, forming distinct clusters. (c) Explicit frame-to-pseudo-gloss alignment on a test sentence: each colored interval marks the contiguous video segment assigned to a predicted content token, with representative frames shown above the timeline.}
    \label{fig:visual}
\end{figure*}

Fig.~\ref{fig:visual}(a) visualizes the learned transport plans on How2Sign: they exhibit clear block-diagonal structure---contiguous frame groups aligning to individual tokens---with the null token absorbing transitional frames, and the ordering is largely monotonic with local reorderings where signing order diverges from the target. Fig.~\ref{fig:visual}(c) marks the video segment assigned to each predicted content token along the timeline, and Fig.~\ref{fig:visual}(b) shows a t-SNE with distinct semantic clusters, indicating alignment that is both temporally coherent at the frame level and semantically organized at the sentence level.

\section{Conclusion}
\label{sec:conclusion}

VTaMo achieves state-of-the-art gloss-free SLT through explicit multi-granularity cross-modal alignment: local OT with a null token, global orthogonal calibration, and position-aligned contrastive learning. Experiments on four benchmarks confirm consistent gains, and ablations verify each component's contribution, showing that explicit alignment is an effective, transferable direction for gloss-free SLT with modest overhead.

\section{Limitations}
\label{sec:limitation}
Our evaluation targets offline accuracy rather than systems-level concerns such as real-time or on-device deployment; the spoken-language order is also recovered by a text-only model rather than observed directly.

\section*{Acknowledgements}
This work was partially supported by ChatSign Technology, Ltd.; and the NYUAD Center for AI and Robotics (CAIR), funded by Tamkeen under the NYUAD Research Institute Award CG010. The generous computational support was provided by the HPC resources at NYU Abu Dhabi and NYU New York.

\bibliographystyle{splncs04}
\bibliography{main}

\newpage
\section*{Appendix}
\addcontentsline{toc}{section}{Appendix}

\setcounter{section}{0}
\setcounter{subsection}{0}
\setcounter{subsubsection}{0}
\setcounter{table}{0}
\setcounter{figure}{0}

\renewcommand{\thesection}{S.\arabic{section}}
\renewcommand{\thesubsection}{S.\arabic{section}.\arabic{subsection}}
\renewcommand{\thesubsubsection}{S.\arabic{section}.\arabic{subsection}.\arabic{subsubsection}}
\renewcommand{\thetable}{S\arabic{table}}
\renewcommand{\thefigure}{S\arabic{figure}}

\noindent
This appendix provides expanded details and additional results complementing the main paper.
Section~\ref{sec:supp_method} presents the full formulations for each component of VTaMo, including the attention-based temporal encoder, the complete local alignment derivation (null-token augmentation, cost matrix, Sinkhorn iterations, epsilon annealing, and loss terms), the global alignment module (attention pooling, memory queue, orthogonal mapping, and EMD loss), the window-based reordering procedure, and the position-aligned contrastive objective.
Section~\ref{sec:supp_experiments} contains supplementary experimental results: a comparison of temporal-pooling strategies (attention pooling vs.\ max-pooling), a comparison against generic video--text alignment objectives (CLIP4Clip, VideoCLIP, X-CLIP), qualitative visualizations of the learned transport plans and t-SNE feature geometry, and representative translation examples on How2Sign comparing VTaMo against SpaMo.

\section{Additional Method Details}
\label{sec:supp_method}

\subsection{Attention-Based Temporal Encoding}
\label{sec:supp_temporal}

The main paper briefly introduces the attention-based temporal encoder.
Here we provide the full formulation.
Prior work commonly uses fixed temporal convolution with max-pooling for downsampling sign language features.
Such pooling discards gradient information from non-selected frames and limits the expressiveness of the learned temporal representation.
We instead use an attention-based temporal convolution module that performs learnable weighted aggregation inside each local temporal region.
The module contains $N_\ell$ cascaded attention-pooling layers.
Each layer performs $2\times$ temporal downsampling, yielding an overall downsampling factor of $2^{N_\ell}$.
Given an input sequence $\mathbf{X} \in \mathbb{R}^{T \times d_h}$, we first apply a one-dimensional convolution with kernel size $5$ and batch normalization.
For each output position $i$, a local window $\mathcal{W}_i$ centered around position $2i$ is extracted.
A learnable query is then constructed from the mean-pooled local context and a positional embedding:
\begin{equation}
\mathbf{q}_i = \frac{1}{|\mathcal{W}_i|}\sum_{j \in \mathcal{W}_i}\mathbf{x}_j + \mathbf{p},
\end{equation}
where $\mathbf{p} \in \mathbb{R}^{d_h}$ denotes the positional embedding.

Multi-head attention with relative position bias is then applied within the window:
\begin{equation}
\alpha_{i,j} =
\mathrm{softmax}
\left(
\frac{\mathbf{q}_i^\top \mathbf{k}_j}{\sqrt{d_h / N_{\mathrm{head}}}}
+
r_{j-c_i}
\right),
\qquad
\mathbf{o}_i = \sum_{j \in \mathcal{W}_i}\alpha_{i,j}\mathbf{v}_j,
\end{equation}
where $c_i$ denotes the center of the local window and $N_{\mathrm{head}}$ is the number of attention heads.
The output is passed through a linear projection, layer normalization, and dropout.
This design allows gradients to flow to all frames in the local region and produces more informative temporal representations than max-pooling.

\subsection{Expanded Local Alignment Formulation}
\label{sec:supp_local_alignment}

The main paper presents the core optimal transport formulation.
This subsection gives the complete construction of the local alignment module.
Let $\mathbf{S} = \{\mathbf{s}_1, \dots, \mathbf{s}_M\} \in \mathbb{R}^{M \times d_t}$ denote the visual features after fusion projection, and let $\mathbf{E} = \{\mathbf{e}_1, \dots, \mathbf{e}_U\} \in \mathbb{R}^{U \times d_t}$ denote the text token embeddings from the frozen encoder embedding layer of the language model.

\subsubsection{Null token augmentation}

Not every visual frame corresponds to a valid lexical unit.
Many frames represent transitional gestures, co-articulation, or short pauses.
To model this explicitly, we introduce a learnable null token $\mathbf{e}_{\varnothing} \in \mathbb{R}^{d_t}$ and prepend it to the text sequence:
\begin{equation}
\tilde{\mathbf{E}} = \{\mathbf{e}_{\varnothing}, \mathbf{e}_1, \dots, \mathbf{e}_U\} \in \mathbb{R}^{K \times d_t},
\qquad
K = U + 1.
\end{equation}
A learnable scalar bias $b_{\varnothing}$ controls the affinity of visual frames to the null token.

\subsubsection{Cost matrix construction}

We compute the pairwise transport cost using cosine distance.
Both modalities are first $\ell_2$ normalized:
\begin{equation}
\bar{\mathbf{s}}_m = \frac{\mathbf{s}_m}{\|\mathbf{s}_m\|_2},
\qquad
\bar{\mathbf{e}}_k = \frac{\tilde{\mathbf{e}}_k}{\|\tilde{\mathbf{e}}_k\|_2}.
\end{equation}
The local cost matrix $\mathbf{C} \in \mathbb{R}^{M \times K}$ is defined as
\begin{equation}
C_{m,k} =
\begin{cases}
1 - \bar{\mathbf{s}}_m^\top \bar{\mathbf{e}}_k - b_{\varnothing}, & \text{if } k = 0, \\
1 - \bar{\mathbf{s}}_m^\top \bar{\mathbf{e}}_k, & \text{otherwise}.
\end{cases}
\label{eq:supp_cost_matrix}
\end{equation}
Columns corresponding to padding tokens are assigned a very large cost to avoid spurious assignments.

\subsubsection{Sinkhorn optimal transport}

We solve the entropy-regularized transport problem
\begin{equation}
\mathbf{A}^{*}
=
\arg\min_{\mathbf{A}\in \Pi(\mathbf{a},\mathbf{b})}
\langle \mathbf{A}, \mathbf{C}\rangle
-
\varepsilon H(\mathbf{A}),
\end{equation}
where $\Pi(\mathbf{a},\mathbf{b})$ denotes the set of transport plans with prescribed marginals, and $H(\mathbf{A}) = -\sum_{m,k} A_{m,k}\log A_{m,k}$ is the entropy term.
The marginals are initialized as uniform distributions over valid visual positions and valid token positions.

For numerical stability, we use the Sinkhorn algorithm in the log domain.
Let $\mathbf{K} = \exp(-\mathbf{C}/\varepsilon)$ denote the Gibbs kernel.
The dual variables are updated as
\begin{align}
\log u_m &\leftarrow \log a_m - \log \sum_k \exp\left(-C_{m,k}/\varepsilon + \log v_k\right), \\
\log v_k &\leftarrow \log b_k - \log \sum_m \exp\left(-C_{m,k}/\varepsilon + \log u_m\right).
\end{align}
After $N_{\mathrm{iter}}$ iterations, the transport plan is recovered as
\begin{equation}
A_{m,k} = \exp\left(\log u_m - C_{m,k}/\varepsilon + \log v_k\right).
\end{equation}

\subsubsection{Epsilon annealing}

To balance exploration and assignment sharpness, we use a multi-stage annealing schedule for $\varepsilon$:
\begin{equation}
\varepsilon(t)=
\begin{cases}
\varepsilon_{\mathrm{high}}, & t \le t_{\mathrm{warm}}, \\
\varepsilon_{\mathrm{high}}
-
\frac{\varepsilon_{\mathrm{high}}-\varepsilon_{\mathrm{mid}}}{t_2-t_{\mathrm{warm}}}
\left(t-t_{\mathrm{warm}}\right), & t_{\mathrm{warm}} < t \le t_2, \\
\varepsilon_{\mathrm{mid}}
-
\frac{\varepsilon_{\mathrm{mid}}-\varepsilon_{\mathrm{low}}}{t_3-t_2}
\left(t-t_2\right), & t_2 < t \le t_3, \\
\varepsilon_{\mathrm{low}}, & t > t_3.
\end{cases}
\label{eq:supp_eps_schedule}
\end{equation}
This schedule keeps the transport plan soft during early training and gradually sharpens the assignments after the alignment becomes stable.

\subsubsection{Local alignment loss terms}

The local alignment objective contains three terms.
The transport cost term is
\begin{equation}
\mathcal{L}_{\mathrm{trans}}
=
\frac{1}{B}
\sum_{i=1}^{B}
\sum_{m,k}
A_{m,k}^{(i)} C_{m,k}^{(i)}.
\label{eq:supp_local_loss}
\end{equation}
To encourage temporal smoothness, we compute a total variation penalty on the row-normalized transport plan restricted to real text tokens.
Let
\begin{equation}
\hat{A}_{m,k}
=
\frac{A_{m,k}}{\sum_{k'=1}^{K-1}A_{m,k'}},
\qquad
k=1,\dots,K-1.
\end{equation}
The temporal variation term is
\begin{equation}
\mathcal{L}_{\mathrm{tv}}
=
\frac{1}{(M-1)(K-1)}
\sum_{m=1}^{M-1}
\sum_{k=1}^{K-1}
\left|
\hat{A}_{m+1,k} - \hat{A}_{m,k}
\right|.
\label{eq:supp_tv_loss}
\end{equation}
To prevent the learnable null bias $b_\varnothing$ from making the null token uniformly cheaper than real tokens, we regulate the cost-based null affinity (computed via row-wise softmax on the cost matrix, independently of the transport plan):
\begin{equation}
p_{\varnothing}^{(m)}
=
\frac{\exp(-C_{m,0}/\tau)}
{\sum_{k=0}^{K-1}\exp(-C_{m,k}/\tau)},
\end{equation}
\begin{equation}
\mathcal{L}_{\mathrm{null}}
=
\left[\bar{p}_{\varnothing}-\rho_{\mathrm{target}}\right]_{+}^{2},
\qquad
\bar{p}_{\varnothing}
=
\frac{1}{M}\sum_{m=1}^{M}p_{\varnothing}^{(m)}.
\label{eq:supp_null_loss}
\end{equation}
The complete local objective is
\begin{equation}
\mathcal{L}_{\mathrm{local}}
=
\beta_{\mathrm{local}} \mathcal{L}_{\mathrm{trans}}
+
\beta_{\mathrm{tv}} \mathcal{L}_{\mathrm{tv}}
+
\beta_{\mathrm{null}} \mathcal{L}_{\mathrm{null}}.
\label{eq:supp_total_local}
\end{equation}

\subsection{Expanded Global Alignment Formulation}
\label{sec:supp_global_alignment}

The main paper introduces the sentence-level global alignment mechanism.
This subsection provides the full formulation.

\subsubsection{Sentence-level attention pooling}

We compute sentence-level visual and textual representations using parameter-free attention pooling.
Given a sequence $\mathbf{X} = \{\mathbf{x}_1,\dots,\mathbf{x}_L\} \in \mathbb{R}^{L \times d_t}$ with validity mask $\mathbf{m} \in \{0,1\}^{L}$, we first define a global query
\begin{equation}
\mathbf{q}
=
\frac{\sum_l m_l \mathbf{x}_l}{\sum_l m_l}.
\end{equation}
The attention weights are then
\begin{equation}
\alpha_l
=
\frac{\exp\left(\mathbf{x}_l^\top \mathbf{q}/\sqrt{d_t}\right)}
{\sum_{l'}\exp\left(\mathbf{x}_{l'}^\top \mathbf{q}/\sqrt{d_t}\right)}.
\end{equation}
The pooled representation is
\begin{equation}
\mathbf{z}
=
\frac{\sum_l \alpha_l \mathbf{x}_l}
{\left\|\sum_l \alpha_l \mathbf{x}_l\right\|_2}.
\label{eq:supp_attn_pool}
\end{equation}
This yields normalized sentence-level vectors $\mathbf{z}^{\mathrm{sign}}$ and $\mathbf{z}^{\mathrm{text}}$ for the visual and textual modalities.

\subsubsection{Memory queue}

The global Earth Mover's Distance objective is most effective when computed over a sufficiently diverse set of sentence pairs.
Since mini-batches are small, we maintain a first-in first-out memory queue that stores recent sentence-level vector pairs.
At each iteration, the current detached sentence vectors are appended to the queue.
The accumulated vectors $\{\mathbf{z}^{\mathrm{sign}}_n\}_{n=1}^{N}$ and $\{\mathbf{z}^{\mathrm{text}}_n\}_{n=1}^{N}$ are then used to compute the global alignment loss.

\subsubsection{Learnable orthogonal mapping}

We introduce a learnable transformation matrix $\mathbf{T} \in \mathbb{R}^{d_t \times d_t}$, initialized near the identity matrix, to map the visual sentence vectors into the textual embedding space:
\begin{equation}
\tilde{\mathbf{z}}^{\mathrm{sign}}_n
=
\frac{\mathbf{z}^{\mathrm{sign}}_n \mathbf{T}}
{\|\mathbf{z}^{\mathrm{sign}}_n \mathbf{T}\|_2}.
\label{eq:supp_global_transform}
\end{equation}

\subsubsection{Global Earth Mover's Distance loss}

Using the transformed visual vectors and the original textual vectors, we define the sentence-level cost matrix
\begin{equation}
C^{g}_{n,n'}
=
1
-
\left(\tilde{\mathbf{z}}^{\mathrm{sign}}_n\right)^\top
\mathbf{z}^{\mathrm{text}}_{n'}.
\end{equation}
A sentence-level transport plan $\mathbf{P}$ is then computed with uniform marginals, and the global alignment loss is
\begin{equation}
\mathcal{L}_{\mathrm{EMD}}
=
\sum_{n,n'} P_{n,n'} C^{g}_{n,n'}.
\label{eq:supp_emd_loss}
\end{equation}

\subsubsection{Orthogonality constraint and scheduled activation}

To preserve pairwise geometry, we regularize the mapping matrix with
\begin{equation}
\mathcal{L}_{\mathrm{orth}}
=
\|\mathbf{T}^{\top}\mathbf{T} - \mathbf{I}\|_{F}^{2}.
\label{eq:supp_orth_loss}
\end{equation}

The global loss is activated only after local alignment becomes reliable.
We use the schedule
\begin{equation}
\lambda_g(t)=
\begin{cases}
0, & t \le t_{\mathrm{warm}}, \\
\lambda_g^{\max}\cdot \frac{t-t_{\mathrm{warm}}}{t_{\mathrm{ramp}}}, & t_{\mathrm{warm}} < t \le t_{\mathrm{warm}} + t_{\mathrm{ramp}}, \\
\lambda_g^{\max}, & t > t_{\mathrm{warm}} + t_{\mathrm{ramp}}.
\end{cases}
\label{eq:supp_lambda_schedule}
\end{equation}
The total global objective is
\begin{equation}
\mathcal{L}_{\mathrm{global}}
=
\lambda_g(t)\mathcal{L}_{\mathrm{EMD}}
+
\beta_{\mathrm{orth}}\mathcal{L}_{\mathrm{orth}}.
\label{eq:supp_total_global}
\end{equation}

\subsection{Window-Based Reordering Guided by Vision--Text Alignment}
\label{sec:supp_reorder}

This section details the window-based reordering module introduced in Sec.~\ref{sec:reordering}. The motivation for reordering is to resolve a fundamental mismatch between the alignment objective and the sequence generation objective. Specifically, the cross-entropy loss supervises the decoder to generate text in the ground-truth word order, whereas the temporal order of visual features does not necessarily follow that same order. This mismatch is particularly pronounced in sign language, where the semantic order expressed in visual signing may differ from the order of the target spoken-language sentence. As a result, if the visual sequence is kept in its original temporal order, the optimization target imposed by cross-entropy can become inconsistent with the correspondence learned by the vision--text alignment module.

To reduce this inconsistency, we reorder the visual feature sequence according to the alignment result before passing it to the decoder. In this way, the reordered visual sequence is made more compatible with the target text order, so that the alignment-induced correspondence and the autoregressive generation objective become mutually consistent during training.

Let $\mathbf{S} = \{\mathbf{s}_1, \dots, \mathbf{s}_M\}$ denote the frame-level visual feature sequence after temporal encoding, and let $\mathbf{A} \in \mathbb{R}^{M \times K}$ denote the local transport plan, where $K = U + 1$ includes the null token and $U$ is the number of text tokens.

Rather than performing reordering at the individual-frame level, we operate on short temporal windows. The reason is that, during inference, the target text order is unknown, and a practically useful alignment pattern should therefore associate a continuous local visual segment with the same text token. If isolated frames were reordered independently, the resulting correspondence could become temporally fragmented, making it difficult for the decoder to recover stable lexical units from the visual stream. Window-based reordering instead encourages locally coherent visual evidence to move together, which better matches the requirement of translation at test time.

We partition the $M$ frame positions into $N_w = U + 2$ overlapping temporal windows. The additional two windows absorb boundary effects. The boundaries of the $w$-th window are defined as
\begin{equation}
s_w = \left\lfloor w \cdot \frac{M-1}{N_w} + 0.5 \right\rfloor,
\qquad
e_w = \left\lfloor (w+1) \cdot \frac{M-1}{N_w} + 0.5 \right\rfloor + 1.
\end{equation}

For each window $w$, we accumulate the alignment mass over the corresponding temporal region:
\begin{equation}
\boldsymbol{\pi}_w
=
\sum_{m=s_w}^{e_w-1}\mathbf{A}_{m,:}
\in \mathbb{R}^{K}.
\end{equation}
The window is then assigned to the real text token with the largest aggregated alignment mass:
\begin{equation}
\phi(w)=\arg\max_{u \in \{1,\dots,U\}} \pi_{w,u}.
\end{equation}
If the dominant mass of a window lies on the null token, that window is excluded from reordering.

For each text token $u$, we collect the set of windows assigned to that token:
\begin{equation}
\Omega_u = \{\, w \mid \phi(w)=u \,\}.
\end{equation}
The reordered visual sequence is then constructed by traversing the text token order from left to right and appending the corresponding windowed visual segments:
\begin{equation}
\mathbf{R}
=
\mathrm{Concat}
\Big(
\{\mathbf{S}_{s_w:e_w-1} \mid w \in \Omega_1^{\uparrow}\},
\{\mathbf{S}_{s_w:e_w-1} \mid w \in \Omega_2^{\uparrow}\},
\dots,
\{\mathbf{S}_{s_w:e_w-1} \mid w \in \Omega_U^{\uparrow}\}
\Big),
\label{eq:supp_direct_window_reorder}
\end{equation}
where $\Omega_u^{\uparrow}$ denotes the windows in $\Omega_u$ sorted by their original temporal order, and $\mathbf{S}_{s_w:e_w-1}$ denotes the original feature subsequence within window $w$.

This design preserves local temporal continuity while rearranging visual segments into an order that is more compatible with the aligned text sequence. Consequently, the reordered representation serves as an intermediate form that better reconciles alignment learning with sequence generation.

\subsection{Expanded Position-Aligned Contrastive Learning}
\label{sec:supp_contrastive}

After reordering, the visual sequence $\mathbf{R} = \{\mathbf{r}_1,\dots,\mathbf{r}_U\}$ is positionally aligned with the text token sequence $\mathbf{E} = \{\mathbf{e}_1,\dots,\mathbf{e}_U\}$.
We use this structure to impose a token-level contrastive objective.
Learnable projection heads are applied to both modalities:
\begin{equation}
\hat{\mathbf{r}}_u
=
\mathrm{LN}
\big(
\mathbf{W}^{v}_2
\sigma(\mathbf{W}^{v}_1 \mathbf{r}_u)
\big),
\qquad
\hat{\mathbf{e}}_u
=
\mathrm{LN}
\big(
\mathbf{W}^{t}_2
\sigma(\mathbf{W}^{t}_1 \mathbf{e}_u)
\big),
\label{eq:supp_proj_heads}
\end{equation}
where $\sigma$ denotes GELU and $\mathrm{LN}$ denotes layer normalization.
After flattening over the batch and valid positions, the InfoNCE loss is
\begin{equation}
\mathcal{L}_{\mathrm{contra}}
=
-
\frac{1}{|\mathcal{V}|}
\sum_{i \in \mathcal{V}}
\log
\frac
{\exp\left(\mathrm{sim}(\hat{\mathbf{r}}_i,\hat{\mathbf{e}}_i)/\tau_c\right)}
{\sum_{j \in \mathcal{V}}\exp\left(\mathrm{sim}(\hat{\mathbf{r}}_i,\hat{\mathbf{e}}_j)/\tau_c\right)},
\label{eq:supp_contrastive_loss}
\end{equation}
where $\mathcal{V}$ denotes the set of valid token positions.
Padding positions are masked out of the denominator.
Gradients from this loss are applied only to the visual branch, while the text branch is detached, so that the pre-trained language model embedding space remains stable.

\section{Additional Qualitative Experiments and Results}
\label{sec:supp_experiments}
\subsection{Attention Pooling versus Max-Pooling}
\label{sec:supp_pooling}
We compare the proposed attention-based temporal convolution module (Sec.~\ref{sec:supp_temporal}) with conventional max-pooling on the Phoenix-2014T~\cite{camgoz2018neural} test set (Table~\ref{tab:pooling}).
Max-pooling keeps only the maximum activation at each position and discards gradient information from all other frames in the pooling window, whereas the attention variant computes a weighted aggregation over all frames, letting gradients reach every frame and yielding more expressive temporal representations.
The attention module improves BLEU-4 from 26.57 to 28.86 and ROUGE-L from 55.23 to 60.24.

\begin{table}[t]
\centering
\caption{Comparison of temporal downsampling strategies on the Phoenix-2014T~\cite{camgoz2018neural} test set.}
\label{tab:pooling}
\begin{tabular}{l c c}
\toprule
Temporal Encoder & BLEU-4 & ROUGE-L \\
\midrule
Conv + Max-Pool & 26.57 & 55.23 \\
Conv + Attention Pool (Ours) & \textbf{28.86} & \textbf{60.24} \\
\bottomrule
\end{tabular}
\end{table}

\subsection{Comparison with Generic Video--Text Alignment}
\label{sec:supp_generic_align}
To test whether the structured correspondence learned by VTaMo could be obtained with off-the-shelf objectives, we replace our three alignment losses with established generic video--text alignment recipes inside the same pipeline (Table~\ref{tab:generic_align}). A CLIP-ViT-only variant with no alignment reaches only 10.45 BLEU-4 on the Phoenix-2014T~\cite{camgoz2018neural} test set, and CLIP4Clip~\cite{luo2022clip4clip}, VideoCLIP~\cite{xu2021videoclip}, and X-CLIP~\cite{ma2022xclip} stay at 9.47--11.13, all far below our 28.86. These recipes align a whole clip to a caption and assume a near-monotonic relation, so they do not resolve the non-monotonic, partial, token-level correspondence that sign language requires.

\begin{table}[t]
\centering
\caption{Plugging generic video--text alignment objectives into our pipeline (Phoenix-2014T~\cite{camgoz2018neural} test set, BLEU-4). Each variant replaces our three alignment losses while keeping all other components fixed.}
\label{tab:generic_align}
\begin{tabular}{l c}
\toprule
Alignment objective & BLEU-4 \\
\midrule
CLIP-ViT only (no alignment)~\cite{radford2021clip} & 10.45 \\
CLIP4Clip~\cite{luo2022clip4clip} & 9.47 \\
VideoCLIP~\cite{xu2021videoclip} & 10.92 \\
X-CLIP~\cite{ma2022xclip} & 11.13 \\
VTaMo (Ours) & \textbf{28.86} \\
\bottomrule
\end{tabular}
\end{table}

\begin{figure*}[t]
\centering
\includegraphics[width=\textwidth]{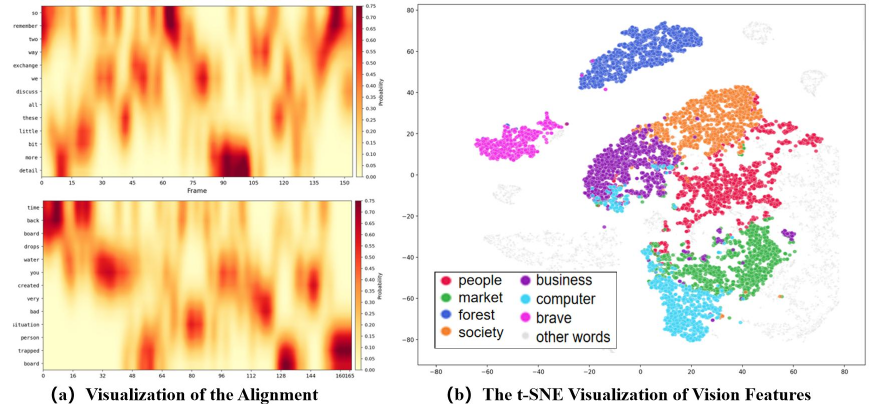}
\caption{Qualitative analysis of the learned alignment and feature geometry. The left panel shows representative transport plans between visual positions and text tokens. The right panel shows a t-SNE visualization of visual features on OpenASL~\cite{shi2022open}.}
\label{fig:visual_analysis}
\end{figure*}

\subsection{Qualitative Analysis of Alignment and Feature Geometry}
\label{sec:supp_qualitative}

We further provide qualitative analysis to examine the behavior of the proposed alignment framework beyond the robustness study. Fig.~\ref{fig:visual_analysis} visualizes the learned transport plans and the resulting feature organization.

The left panel shows representative transport plans between temporal visual positions and text tokens. The alignment maps exhibit clear structured correspondences, indicating that contiguous temporal regions are consistently associated with semantically relevant target tokens. The right panel shows a t-SNE visualization of the resulting visual features. Semantically related samples form compact groups, which suggests that the alignment objectives induce a more organized cross-modal representation space.

\subsection{Representative Translation Examples}
\label{sec:supp_translation}

Table~\ref{tab:supp_translation_examples} presents representative translation examples on the How2Sign~\cite{duarte2021how2sign} test set.
Compared with SpaMo~\cite{hwang2025efficient}, our method more faithfully preserves key content words and overall sentence meaning.
The advantage is especially clear in examples that require precise lexical grounding, attribute preservation, and procedural understanding.

\begin{table}[t]
\centering
\caption{Representative translation examples on the How2Sign~\cite{duarte2021how2sign} test set, comparing SpaMo~\cite{hwang2025efficient} and Ours on five examples.}
\label{tab:supp_translation_examples}
\resizebox{\linewidth}{!}{
\begin{tabular}{l p{10.8cm}}
\toprule
Source & Text \\
\midrule
Reference & And everyone knows a quick bat is what makes you effective. \\
SpaMo & You know that this motion is fast. \\
Ours & Everyone knows a fast bat can be effective.  \\
\cmidrule(lr){1-2}
Reference & In this way, we know that we're taking something that is legal and properly prescribed for us as an individual.
 \\
SpaMo & Now we know this thing is good for a person but we should not share it. \\
Ours & We know that we drink something that is right and prescribed for us as a single one. \\
\cmidrule(lr){1-2}
Reference & And we all ultimately want to be the big guy that trains a lot of other people so it's kind of a historical tradition to mentor and be mentored. \\
SpaMo & We want to became bigger and walk with many people and this will be happened in the office with your teacher. \\
Ours & We all finally want to be the big man that trains many other people, so it's a historical tradition to mentor and be educated. \\
\cmidrule(lr){1-2}
Reference &  So before you begin to work on your horse there are a number of things that you need to keep in mind \\
SpaMo & You should go to look for your hat when you riding the horse. \\
Ours & When you work on the horse there are some things that you should remember. \\
\cmidrule(lr){1-2}
Reference & We'll talk about the theory and what that means later but it allow you to play an open chord without putting any hands on the guitar.
 \\
SpaMo & We find that this class has a lot of things to talk but I plan to play with my friend rather than involving it. \\
Ours & We talk about the theory and what it means, but it makes you to play the guitar without putting your hands. \\
\bottomrule
\end{tabular}}
\end{table}

\end{document}